\documentclass{article}

\usepackage{PRIMEarxiv}
\usepackage[utf8]{inputenc} 
\usepackage[T1]{fontenc}    
\usepackage{hyperref}       
\usepackage{url}            
\usepackage{booktabs}       
\usepackage{amsmath,amssymb,amsfonts}       
\usepackage{nicefrac}       
\usepackage{microtype}      
\usepackage{lipsum}
\usepackage{fancyhdr}       
\usepackage{graphicx}       
\graphicspath{{media/}}     
\usepackage{multirow}
\usepackage{subcaption}
\usepackage{threeparttable}

\title{The Power of Combining Data and Knowledge: GPT-4o is an Effective Interpreter of Machine Learning Models in Predicting Lymph Node Metastasis of Lung Cancer
\thanks{\textit{\underline{Citation}}: 
\textbf{Authors. Title. Pages.... DOI:000000/11111.}} 
}

\author{
  Danqing Hu \\
  Zhejiang Lab \\
  Hangzhou, Zhejiang, China\\
  \texttt{hudq@zhejianglab.com} \\
   \And
  Bing Liu\\
  Peking University Cancer Hospital and Institute \\
  Beijing, China\\
  \texttt{liubing983811735@126.com} \\  
   \And
  Xiaofeng Zhu \\
  Zhejiang Lab \\
  Hangzhou, Zhejiang, China\\
  \texttt{andy.zhu@zhejianglab.com} \\
  \And
  Nan Wu \\
  Peking University Cancer Hospital and Institute \\
  Beijing, China\\
  \texttt{nanwu@bjmu.edu.cn} \\
}

\begin{document}
\maketitle

\begin{abstract}

Lymph node metastasis (LNM) is a crucial factor in determining the initial treatment for patients with lung cancer, yet accurate preoperative diagnosis of LNM remains challenging. Recently, large language models (LLMs) have garnered significant attention due to their remarkable text generation capabilities. Leveraging the extensive medical knowledge learned from vast corpora, LLMs can estimate probabilities for clinical problems, though their performance has historically been inferior to data-driven machine learning models. In this paper, we propose a novel ensemble method that combines the medical knowledge acquired by LLMs with the latent patterns identified by machine learning models to enhance LNM prediction performance. Initially, we developed machine learning models using patient data. We then designed a prompt template to integrate the patient data with the predicted probability from the machine learning model. Subsequently, we instructed GPT-4o, the most advanced LLM developed by OpenAI, to estimate the likelihood of LNM based on patient data and then adjust the estimate using the machine learning output. Finally, we collected three outputs from the GPT-4o using the same prompt and ensembled these results as the final prediction. Using the proposed method, our models achieved an AUC value of 0.778 and an AP value of 0.426 for LNM prediction, significantly improving predictive performance compared to baseline machine learning models. The experimental results indicate that GPT-4o can effectively leverage its medical knowledge and the probabilities predicted by machine learning models to achieve more accurate LNM predictions. These findings demonstrate that LLMs can perform well in clinical risk prediction tasks, offering a new paradigm for integrating medical knowledge and patient data in clinical predictions.

\end{abstract}


\section{Introduction}

Lung cancer remains the leading cause of cancer-related mortality globally \cite{Sung2021}. For patients with early-stage lung cancer, surgical resection represents the only potentially curative treatment \cite{Howington2013}. The determination of lymph node metastasis (LNM) is critical in assessing surgical eligibility and the need for additional neoadjuvant therapy. However, accurately diagnosing LNM preoperatively through non-invasive examinations and tests poses significant challenges in clinical practice, often leading to suboptimal treatment decisions and adversely affecting patient outcomes \cite{Navani2018}.

To achieve accurate preoperative diagnosis of LNM, data-driven approaches have become the most employed methods for developing LNM prediction models. Initially, researchers utilized patients' clinical features in combination with statistical methods to construct predictive models \cite{Farjah2013,Chen2013}. To leverage imaging data, the radiomics approach was introduced, allowing the extraction of first-order, second-order, texture, and other features from image data, which were then integrated with clinical features to enhance predictive accuracy \cite{Gu2018,He2019,Wang2019}.

To further explore the nonlinear relationships among these features, machine learning methods such as random forest, support vector machine, and multilayer perceptron were employed, resulting in improved model performance \cite{Wang2018,Cong2020,Yoo2020,Hu2022}. With the rapid advancement of deep learning, researchers began using deep learning techniques to automatically extract deep features from images for LNM prediction \cite{Zhao2020,Wang2017,Wang2021,Hu2023,Hu2024}. Unlike radiomics methods, deep learning approaches do not require manual delineation of regions of interest in the images. Instead, they can directly extract deep image features related to the prediction target through error backpropagation, making deep learning the most popular and effective approach for LNM prediction.

Recently, large language models (LLMs), such as ChatGPT \cite{OpenAI2024} and GPT-4 \cite{Achiam2023}, have captured global attention due to their impressive text-generation capabilities. These models, pre-trained on vast corpora, demonstrate remarkable performance on previously unseen tasks using zero-shot, one-shot, or few-shot prompts without parameter updates \cite{Brown2020}. By incorporating reinforcement learning from human feedback (RLHF) \cite{Ouyang2022}, LLMs are further refined to produce content that is safe and aligns with human expectations. This success has led to a paradigm shift in natural language processing (NLP) research and is gradually influencing clinical prediction research \cite{Tang2023,HuChen2024,Doshi2024,HuZhang2024,HuLiu2023}.

Leveraging the medical knowledge learned from extensive training data, LLMs show potential in diagnosing and evaluating patient prognoses. Many studies have investigated the capabilities of LLMs in predicting clinical outcomes such as readmission, length of stay, and hospital mortality \cite{Chung2024,Glicksberg2024,Changho2024,ZhuWang2024}. These studies typically develop prompts using patient data and instruct the LLMs to provide answers for specific tasks. While LLMs can generate predictive results based on prompted patient information and instructions, their predictive performance rarely surpassed that of traditional data-driven machine learning models \cite{Chung2024,Glicksberg2024}.

In this study, we propose a novel method that integrates the medical knowledge of LLMs with the latent patterns identified by data-driven models to predict lymph node metastasis in lung cancer. Our approach demonstrates that by combining the strengths of both knowledge-based and data-driven models, we can achieve superior predictive performance compared to using either model alone.

\section{Materials and methods}

\subsection{Patients}

We collected data from 767 lung cancer patients treated at Peking University Cancer Hospital. All patients underwent pulmonary resection with systematic mediastinal lymphadenectomy between 2010 and 2018 and received contrast-enhanced computed tomography (CT) scans and tumor biomarker tests within two months before surgery. Patients who received preoperative chemotherapy or radiotherapy were excluded to avoid potential confounding from complete responses to these treatments.

The collected data included structured clinical information such as demographics, tumor biomarkers as well as unstructured data like disease history, CT and pathological reports. One clinician annotated lymph node metastasis (LNM) statuses based on post-operative pathological reports, which served as the gold standard labels. Ethical approval for this study was granted by the Ethics Committee of Peking University Cancer Hospital (2022KT128) prior to this study.

\begin{figure}[h]
\centering
\includegraphics[width=\textwidth]{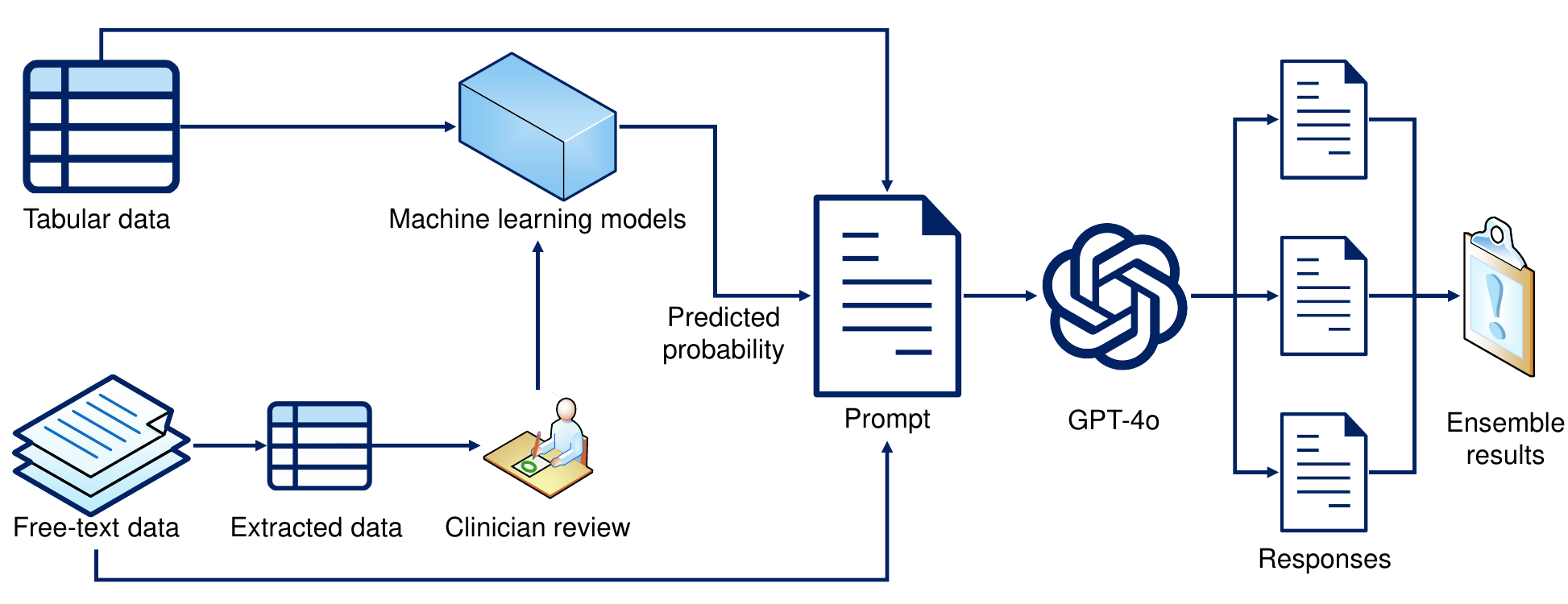}
\caption{Overall study design.}
\label{figure1}
\end{figure}

\subsection{Study design}

This study aims to integrate the advantages of large language models (LLMs) and machine learning models to accurately predict lymph node metastasis in lung cancer patients. The overall study design is depicted in Figure \ref{figure1}.

First, unstructured clinical data were collected and key features were extracted using information extraction models previously developed by our team \cite{HuLiu2023,HuZhangLi2021}. These extracted features were then reviewed by clinicians. Next, we developed machine learning models using the reviewed structured features from training set to predict the risk of lymph node metastasis of patients in the test set. We then constructed prompts for GPT-4o using the predicted probabilities and patient information, and gathered several responses from GPT-4o using the same prompt. Finally, we integrated the various predicted results from GPT-4o to generate the final ensemble results.

\subsection{Machine learning models}

In this study, we select three classical machine learning methods, i.e., logistic regression, random forest, and support vector machine, to identify latent patterns between patient clinical data and LNM status. A 10-fold cross-validation strategy was employed for training and testing the models. During each fold iteration, we utilized an additional 5-fold cross-validation to optimize hyperparameters, subsequently retraining the model on the entire training set using the best hyperparameters. The trained model was then tested on the test set to obtain the final test results. After completing all 10-fold iterations, we obtained 10 test results for each fold and the predicted probability of LNM for each patient. We will use the test results and predicted probabilities to construct the prompt to achieve the integration of data and knowledge.

\begin{figure}[h]
\centering
\includegraphics[width=0.5\textwidth]{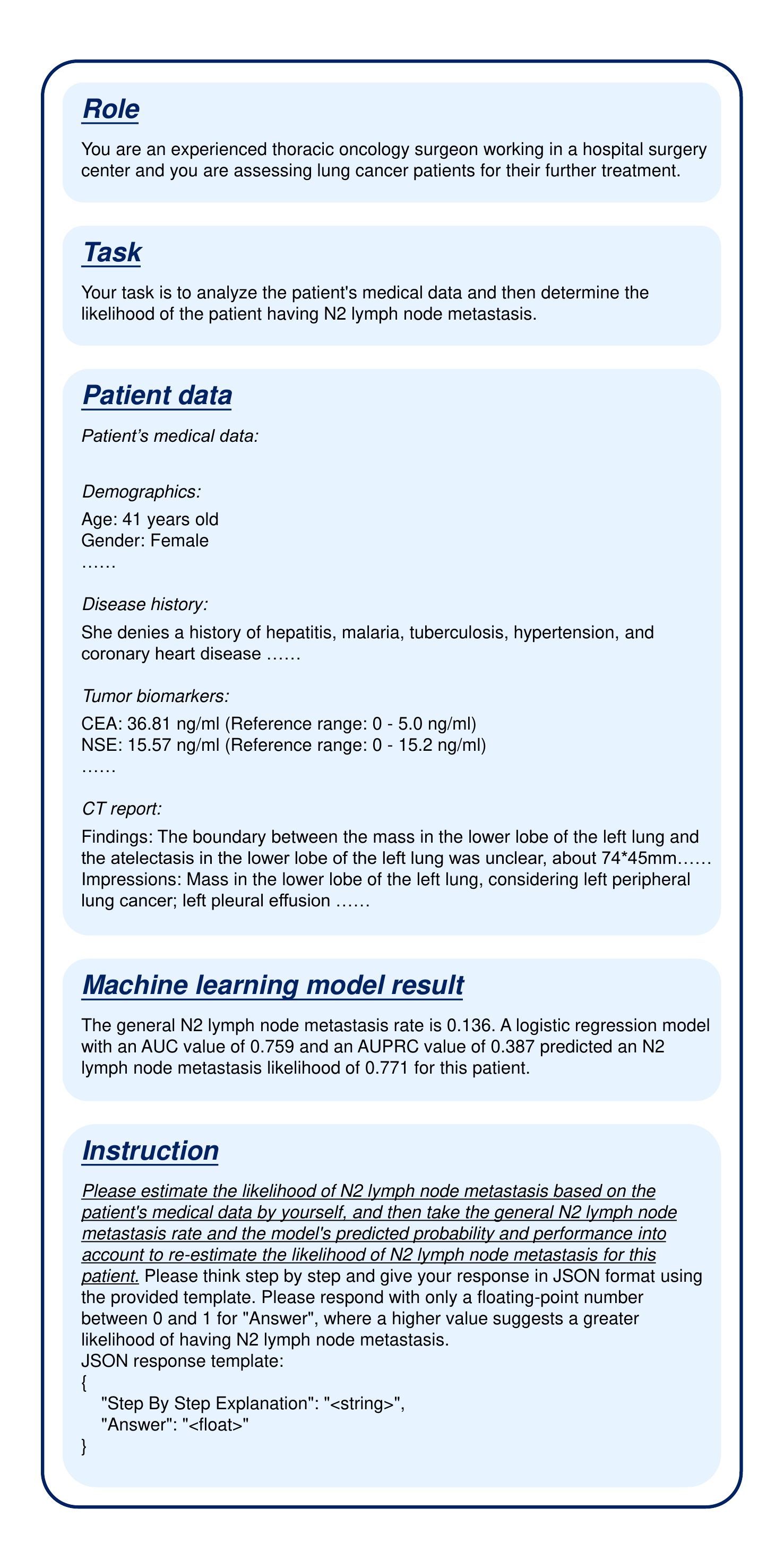}
\caption{Prompt template.}
\label{figure2}
\end{figure}

\subsection{Prompt design}

The prompt template used in this study is shown in Figure \ref{figure2}. It consists of 5 elements, i.e., Role, Task, Patient data, machine learning model result, Instruction.

\begin{itemize}
\item \textbf{Role} This element defines the role that LLMs should assume to generate responses for specific tasks. In this study, we instructed the LLMs to act as thoracic surgeons, who typically assess a patient's LNM and determine whether the patient can receive surgical resection directly.

\item \textbf{Task} This element specifies the clinical prediction task assigned to the LLMs. We instructed the LLMs to predict the likelihood of a patient having N2 lymph node metastasis.

\item \textbf{Patient data} This element outlines the patient clinical data used for the evaluation by the large language models (LLMs). We provided patient demographics, disease history, tumor biomarkers, and CT reports. It is important to note that the original disease history and CT reports were in Chinese free-text format, therefore, we used the Google Translate API via googletrans to translate them into English. Additionally, for tumor biomarkers, we supplied the reference ranges as external knowledge.

\item \textbf{Machine learning model result} This element is used to integrate the predicted result from the data-driven model as a reference for the LLMs. In addition to the predicted probability, we included the type of model and its predictive performance (AUC value and AP value). Furthermore, we provided the N2 LNM rate among the sample to support the LLMs' evaluation.

\item \textbf{Instruction} In this element, we instructed the LLMs to initially estimate the likelihood of N2 LNM based solely on the patient data. Subsequently, they were to re-estimate the likelihood by considering the N2 LNM rate and the predicted probability and performance of the machine learning model. We also employed the Chain-of-Thought strategy to require the LLMs to reason step by step. Additionally, the LLMs were instructed to provide their responses in JSON format with key-value pairs, such as "Step By Step Explanation":"<string>" and "Answer":"<float>".

\end{itemize}

\subsection{Ensemble models}

Using the designed prompt template, we developed individualized prompts for each patient sample and utilized the OpenAI API to obtain responses. We selected OpenAI's most advanced model, GPT-4o (gpt-4o-2024-05-13), to generate these responses.

Considering that large language models (LLMs) can produce varying outputs even with identical prompts, we input the same prompt three times for each patient to obtain three distinct responses. We then applied four strategies, i.e., maximum value, minimum value, median value, and mean value, to process these three responses and derive the ensemble results.

\subsection{Experimental setup}

To compare with the proposed method, we selected GPT-3.5 (ChatGPT) as the baseline model. Additionally, we excluded the "Machine Learning Model Results" element to assess the performance of LLMs alone for predicting N2 lymph node metastasis (LNM). The prompt template used for GPT-3.5 and GPT-4 is provided in Figure \ref{figure3} in Supplement.

We further redesigned the prompt template to remove the instruction about estimating the likelihood of N2 lymph node metastasis (LNM) using patient data first. This modification allowed us to explore how the LLMs utilize patient data and predicted probabilities to generate their own results. The redesigned prompt template is provided in Figure \ref{figure4} in Supplement.

Model performance was evaluated using two metrics: the area under the receiver operating characteristic curve (AUC) and the average precision value (AP).

To test the differences in performance between models, we employed the paired t-test. A p-value of less than 0.05 was considered statistically significant.

\section{Results}

\subsection{Clinical data}

Among the 767 patients, 104 (13.6\%) were confirmed to have N2 LNM according to their postoperative pathology reports. A total of 26 types of clinical features were included in this study. Features such as spiculation, lobulation, mediastinal lymph node short axis, hilar lymph node short axis, tumor location, and tumor density were extracted from the CT reports and reviewed by a clinician. Table \ref{table1} presents the statistics of the clinical data.

\begin{table}[t]
    \caption{The statistics of the clinical data.}
    \setlength{\tabcolsep}{1pt}
    \centering
    \begin{tabular}{lcc|lcc}
    \toprule
         Clinical feature & Positive (n=104) & Negative (n=663) & Clinical feature & Positive (n=104) & Negative (n=663) \\
    \midrule
         Age                             & 60.82  $\pm$ 9.02   & 60.79  $\pm$ 9.53   & Cardiovascular diseases          &         &      \\
         Height                          & 164.57 $\pm$ 6.93   & 164.50 $\pm$ 7.92   & \ \ \ \ Yes                      & 9       & 27   \\
         Weight                          & 66.93  $\pm$ 9.47   & 65.59  $\pm$ 9.50   & \ \ \ \ No                       & 95      & 636  \\
         Tumor long size                 & 3.01   $\pm$ 1.38   & 2.56   $\pm$ 1.40   & Cerebrovascular diseases         &         &      \\
         Tumor short size                & 2.38   $\pm$ 1.11   & 1.99   $\pm$ 1.16   & \ \ \ \ Yes                      & 6       & 23   \\
         CEA                             & 12.76  $\pm$ 21.18  & 4.24   $\pm$ 9.53   & \ \ \ \ No                       & 98      & 640  \\
         CA199                           & 15.89  $\pm$ 20.96  & 13.95  $\pm$ 15.39  & Spiculation                      &         &      \\
         CA125                           & 19.96  $\pm$ 25.55  & 13.47  $\pm$ 10.18  & \ \ \ \ Yes                      & 39      & 171  \\
         NSE                             & 16.25  $\pm$ 6.10   & 15.68  $\pm$ 7.02   & \ \ \ \ No                       & 65      & 492  \\
         CYFRA211                        & 3.57   $\pm$ 4.21   & 3.18   $\pm$ 3.34   & Lobulation                       &         &      \\
         SCCAG                           & 1.19   $\pm$ 1.81   & 0.93   $\pm$ 0.97   & \ \ \ \ Yes                      & 52      & 174  \\
         Gender                          &  &                                        & \ \ \ \ No                       & 52      & 489  \\
         \ \ \ \ Male                    & 62                  & 322                 & MLNSA                            &         &      \\
         \ \ \ \ Female                  & 42                  & 341                 & \ \ \ \ $\geq$10mm               & 34      & 80   \\
         Smoking history                 &  &                                        & \ \ \ \ $<$10mm                  & 70      & 583  \\
         \ \ \ \ Yes                     & 55                  & 272                 & HLNSA                            &         &      \\
         \ \ \ \ No                      & 49                  & 391                 & \ \ \ \ $\geq$10mm               & 23      & 71   \\
         Drinking history                &  &                                        & \ \ \ \ $<$10mm                  & 81      & 592  \\
         \ \ \ \ Yes                     & 25                  & 151                 & Tumor location                   &         &      \\
         \ \ \ \ No                      & 79                  & 512                 & \ \ \ \ RUL                      & 27      & 209  \\
         Family tumor history            &  &                                        & \ \ \ \ RML                      & 4       & 54   \\
         \ \ \ \ Yes                     & 14                  & 116                 & \ \ \ \ RLL                      & 18      & 129  \\
         \ \ \ \ No                      & 90                  & 547                 & \ \ \ \ LUL                      & 27      & 140  \\ 
         Hypertension                    &  &                                        & \ \ \ \ LLL                      & 21      & 100  \\
         \ \ \ \ Yes                     & 37                  & 184                 & \ \ \ \ Others                   & 7       & 31   \\
         \ \ \ \ No                      & 67                  & 479                 & Tumor density                    &         &      \\
         Diabetes                        &  &                                        & \ \ \ \ Solid                    & 101     & 457  \\
         \ \ \ \ Yes                     & 14                  & 65                  & \ \ \ \ mGGO                     & 3       & 92   \\
         \ \ \ \ No                      & 90                  & 598                 & \ \ \ \ GGO                      & 0       & 114  \\
         Tuberculosis history            &  &                                        &  &  &   \\
         \ \ \ \ Yes                     & 2                   & 29                  &  &  &   \\
         \ \ \ \ No                      & 102                 & 634                 &  &  &   \\
    \bottomrule
    \multicolumn{6}{p{460pt}}{CEA: carcinoembryonic antigen, CA199: carbohydrate antigen 19-9, CA125: carbohydrate antigen 12-5, NSE: neuron-specific enolase, Cyfra211: cytokeratin 19-fragments, SCCAg: squamous cell carcinoma antigen, MLNLA: mediastinal lymph node long axis, MLNSA: mediastinal lymph node short axis, RUL: right upper lobe, RML: right middle lobe, RLL: right lower lobe, LUL: left upper lobe, LLL: left lower lobe.}\\
    \end{tabular}
    
    \label{table1}
\end{table}

\subsection{Predictive performance}

Table \ref{table2} presents the predictive performance of the baseline and proposed models. GPT-4o consistently outperforms GPT-3.5 in terms of both AUC and AP values, indicating a superior capability for LNM prediction. However, the plain GPT-4o model does not surpass the performance of the ML models, suggesting that data-driven models can extract more valuable latent patterns from patient data than those learned by medical knowledge-based LLMs for predicting LNM risk.

A comparison of the ML models with GPT-4o+ML models reveals that their performances are similar. However, when applying the proposed ensemble method, the ensemble models outperform the ML models, particularly with the mean, median, and min ensemble strategies. These findings indicate that LLMs can leverage their medical knowledge to estimate LNM likelihood using patient data, and then refine the results by incorporating the predictions of ML models. This highlights the advantage of integrating data-driven and knowledge-based approaches.

\begin{table}
    \centering
    \caption{The AUC and AP values of the baseline and proposed models.}
    \setlength{\tabcolsep}{12pt}
    \begin{tabular}{lcccc}
    \toprule
    \multirow{2}{*}{Models} & \multicolumn{2}{l}{AUC}       & \multicolumn{2}{l}{AP}\\
                            & Mean       & SD               & Mean   & SD            \\ 
    \midrule
    GPT-3.5                 & 0.687	& 0.054	& 0.242	& 0.035\\
    GPT-4o                  & 0.714	& 0.083	& 0.284	& 0.074\\
    \hline
    LR                      & 0.759	& 0.036	& 0.387	& 0.075\\
    GPT-4o+LR*              & 0.759	& 0.035	& 0.387	& 0.074\\
    GPT-4o+LR max           & 0.762	& 0.036	& 0.391	& 0.071\\
    GPT-4o+LR min           & 0.765	& 0.046	& 0.417	& 0.081\\
    \textbf{GPT-4o+LR median}        & \textbf{0.770}	& 0.039	& 0.403	& 0.089\\
    \textbf{GPT-4o+LR mean}          & 0.767	& 0.039	& \textbf{0.423}	& 0.084\\
    \hline
    RF                      & 0.752	& 0.054	& 0.402	& 0.107\\
    GPT-4o+RF*              & 0.749	& 0.060	& 0.400	& 0.110\\
    GPT-4o+RF max           & 0.768	& 0.057	& 0.390	& 0.102\\
    GPT-4o+RF min           & 0.776	& 0.055	& 0.384	& 0.094\\
    GPT-4o+RF median        & 0.775	& 0.064	& 0.412	& 0.112\\
    \textbf{GPT-4o+RF mean}          & \textbf{0.778}	& 0.058	& \textbf{0.426}	& 0.106\\
    \hline
    SVM                     & 0.749	& 0.031	& 0.379	& 0.063\\
    GPT-4o+SVM*             & 0.744	& 0.031	& 0.365	& 0.065\\
    GPT-4o+SVM max          & 0.755	& 0.036	& 0.360	& 0.076\\
    \textbf{GPT-4o+SVM min}          & 0.760	& 0.030	& \textbf{0.383}	& 0.063\\
    \textbf{GPT-4o+SVM median}       & \textbf{0.764}	& 0.031	& 0.375	& 0.061\\
    GPT-4o+SVM mean         & 0.758	& 0.030	& 0.380	& 0.062\\
    \bottomrule
    \multicolumn{5}{p{200pt}}{*: models using baseline prompt template II}\\
    \end{tabular}
    \label{table2}
\end{table}

Table \ref{table3} presents a statistical analysis comparing the performance of ML models with ensemble GPT-4o+ML models. The results reveal that the mean ensemble strategy (GPT-4o+ML mean) significantly improves the AUC and AP values of the LR model and the AUC values of the RF model. Similarly, the median ensemble strategy (GPT-4o+ML median) significantly improves the AUC values of the LR, RF, and SVM models. Besides, the max ensemble strategy (GPT-4o+ML max) improves the AUC value of the RF model significantly. Although the min ensemble strategy (GPT-4o+ML min) achieves competitive AUC and AP values compared with the ML models, the differences are not statistically significant. Based on the statistical analysis, we can note that the mean and median ensemble strategies show better results than the max and min ensemble strategies.

\begin{table}
    \centering
    \caption{The statistical analysis of the performance of ML models and ensemble GPT-4o+ML models.}
    
    \begin{tabular}{lcc}
    \toprule
    Models                  & p-value (AUC) & p-value (AP) \\
    \midrule
    LR vs GPT-4o+LR*        & 0.360          & 0.198\\
    LR vs GPT-4o+LR max     & 0.495          & 0.752\\
    LR vs GPT-4o+LR min     & 0.252          & 0.063\\
    \textbf{LR vs GPT-4o+LR median}  & \textbf{0.015}          & 0.316\\
    \textbf{LR vs GPT-4o+LR mean}    & \textbf{0.033}          & \textbf{0.043}\\
    \hline
    RF vs GPT-4o+RF*        & 0.361          & 0.067\\
    \textbf{RF vs GPT-4o+RF max}     & \textbf{0.048}          & 0.494\\
    RF vs GPT-4o+RF min     & 0.100          & 0.529\\
    \textbf{RF vs GPT-4o+RF median}  & \textbf{0.020}          & 0.614\\
    \textbf{RF vs GPT-4o+RF mean}    & \textbf{0.016}          & 0.216\\
    \hline
    SVM vs GPT-4o+SVM*        & 0.066          & 0.101\\
    SVM vs GPT-4o+SVM max     & 0.636          & 0.319\\
    SVM vs GPT-4o+SVM min     & 0.061          & 0.725\\
    \textbf{SVM vs GPT-4o+SVM median}  & \textbf{0.023}          & 0.719\\
    SVM vs GPT-4o+SVM mean    & 0.297          & 0.943\\
    \bottomrule
    \multicolumn{3}{p{200pt}}{*: models using baseline prompt template II}\\
    \end{tabular}
    \label{table3}
\end{table}

\subsection{LLMs' capabilities of using ML models}

In addition to GPT-4o, we also evaluated GPT-3.5 (ChatGPT) as a baseline to investigate the capabilities of LLMs in using machine learning models' results to predict LNM. The experimental results for GPT-3.5 are presented in Table \ref{table4}. It is evident that GPT-3.5 did not outperform the machine learning models, even when provided with the predicted probabilities from these models. This finding emphasizes the importance of advanced reasoning abilities and extensive medical knowledge in LLMs for clinical prediction tasks.

\begin{table}
    \centering
    \caption{The AUC and AP values of the GPT-3.5+ML models.}
    \setlength{\tabcolsep}{12pt}
    \begin{tabular}{lcccc}
    \toprule
    \multirow{2}{*}{Models}  & \multicolumn{2}{l}{AUC}       & \multicolumn{2}{l}{AP}\\
                             & Mean       & SD               & Mean   & SD            \\ 
    \midrule
    GPT-3.5                  & 0.687	& 0.054	& 0.242	& 0.035\\
    \hline
    \textbf{LR}                       & \textbf{0.759}	& 0.036	& \textbf{0.387}	& 0.075\\
    GPT-3.5+LR max           & 0.753	& 0.044	& 0.377	& 0.086\\
    GPT-3.5+LR min           & 0.750	& 0.038	& 0.378	& 0.076\\
    GPT-3.5+LR median        & 0.750	& 0.042	& 0.376	& 0.082\\
    GPT-3.5+LR mean          & 0.757	& 0.037	& 0.383	& 0.076\\
    \hline
    \textbf{RF}                       & 0.752	& 0.054	& \textbf{0.402}	& 0.107\\
    GPT-3.5+RF max           & 0.725	& 0.048	& 0.274	& 0.053\\
    GPT-3.5+RF min           & 0.747	& 0.084	& 0.349	& 0.103\\
    GPT-3.5+RF median        & 0.749	& 0.056	& 0.330	& 0.079\\
    \textbf{GPT-3.5+RF mean}          & \textbf{0.753}	& 0.060	& 0.362	& 0.084\\
    \hline
    \textbf{SVM}                      & \textbf{0.749}	& 0.031	& \textbf{0.379}	& 0.063\\
    GPT-3.5+SVM max          & 0.694	& 0.047	& 0.273	& 0.067\\
    GPT-3.5+SVM min          & 0.736	& 0.052	& 0.363	& 0.074\\
    GPT-3.5+SVM median       & 0.721	& 0.038	& 0.317	& 0.076\\
    GPT-3.5+SVM mean         & 0.722	& 0.046	& 0.327	& 0.072\\
    \bottomrule
    \end{tabular}
    \label{table4}
\end{table}

\section{Discussion}

In this study, we proposed a novel ensemble approach to instruct large language models (LLMs) to leverage the predicted likelihoods from data-driven models alongside their own medical knowledge to achieve better predictive performance for lymph node metastasis (LNM) prediction. The powerful capabilities of LLMs suggest a new paradigm for integrating patient data and medical knowledge to predict clinical risk.

One critical factor in achieving improved performance with LLMs is instructing them to estimate risk based on patient data first and then adjust these results using the predicted outcomes from machine learning models. This instruction is essential for LLMs' reasoning, as baseline models without this directive did not show performance improvements. Step-by-step explanations of the baseline model indicate that LLMs tend to default to the predicted likelihoods of machine learning models if not explicitly guided to estimate independently.

The performance improvements varied depending on the type of machine learning model used. For instance, when using a logistic regression model, GPT-4 primarily improved the AUC and AP value. In contrast, for RF and SVM models, GPT-4 mainly enhanced the AUC values. These variations arise because different machine learning models produce different predicted probabilities for the same patient, and the LLM adjusts its predictions based on these probabilities.

However, the current study has some limitations. First, the evaluation of the proposed approach was limited to the LNM prediction task. The medical knowledge that LLMs acquire may differ across various diseases and clinical problems. Future research should explore whether LLMs can achieve better predictive performance for other clinical prediction tasks.

Additionally, this study did not incorporate image data to create a multimodal prediction task. Some studies have explored using LLMs like GPT-4 to diagnose diseases using image data. However, they did not show competitive performance in interpreting real-world medial image \cite{YanZhang2023,Nakao2024,ZhouOng2024,Brin2024}. Future research should investigate how to integrate image data to enhance predictive performance further.

\section{Conclusion}

In this study, we proposed a novel approach for LNM prediction by leveraging the medical knowledge acquired by large LLMs and the latent patterns learned by machine learning models. Our experimental results demonstrate that GPT-4o can effectively adjust its own predicted risks based on machine learning model predictions, achieving significantly improved performance. These findings suggest that LLMs can excel in clinical risk prediction tasks, offering a new paradigm for integrating medical knowledge and patient data in clinical predictions.

\section*{Acknowledgments}

This work was supported by the Beijing Natural Science Foundation (L222020), the National Key R\&D Program of China (No.2022YFC2406804), the Capital’s funds for health improvement and research (No.2024-1-1023), and the National Ten-thousand Talent Program.

\section*{Supplements}

\begin{figure}[h]
\centering
\includegraphics[width=0.5\textwidth]{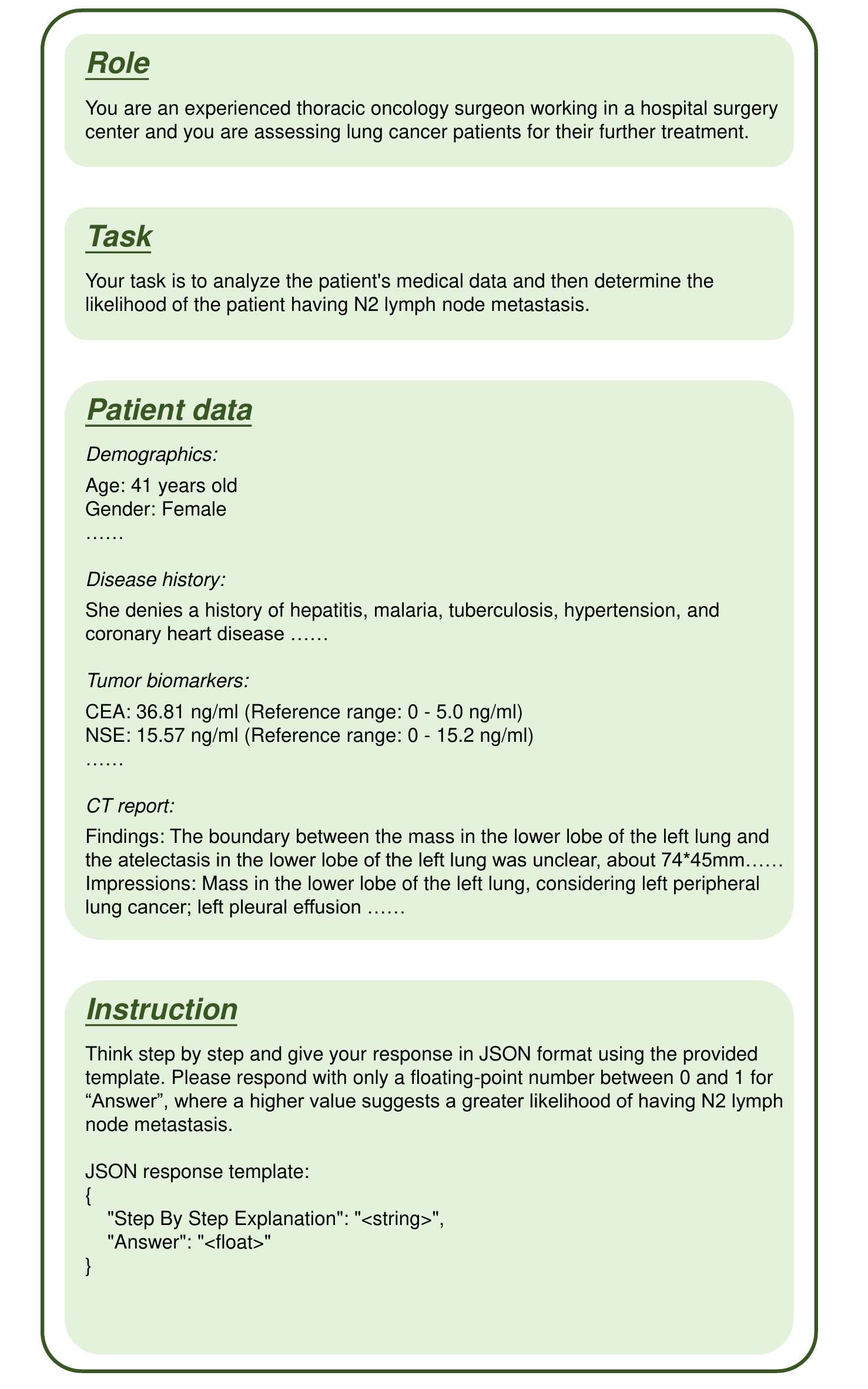}
\caption{Baseline prompt template I.}
\label{figure3}
\end{figure}

\begin{figure}[h]
\centering
\includegraphics[width=0.5\textwidth]{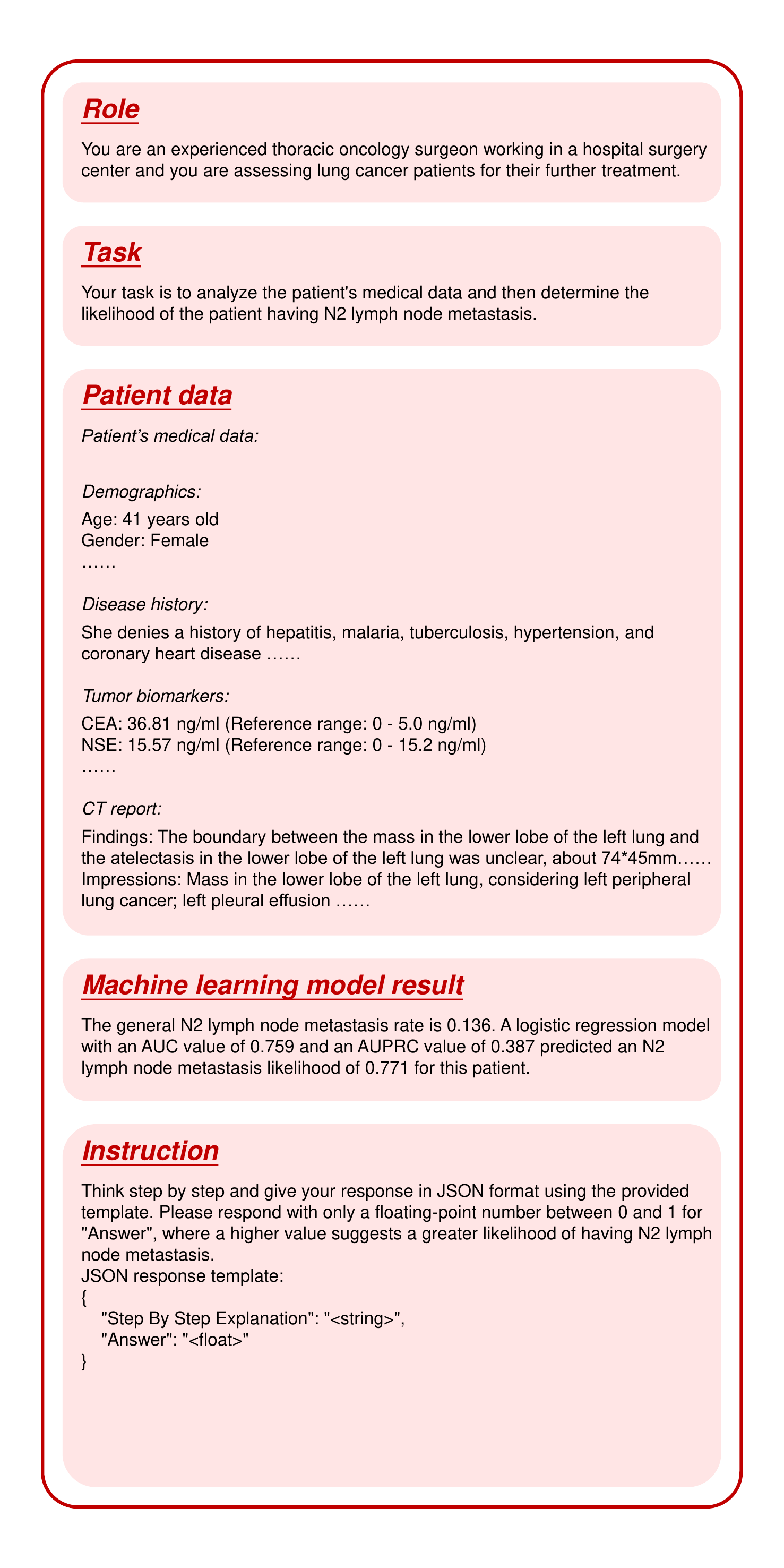}
\caption{Baseline prompt template II.}
\label{figure4}
\end{figure}


\begin{thebibliography}{40}

\bibitem{Sung2021} H. Sung et al., "Global Cancer Statistics 2020: GLOBOCAN Estimates of Incidence and Mortality Worldwide for 36 Cancers in 185 Countries," CA: A Cancer Journal for Clinicians, vol. 71, no. 3, pp. 209-249, 2021.



 
\bibitem{Howington2013} J. A. Howington, M. G. Blum, A. C. Chang, A. A. Balekian, and S. C. Murthy, "Treatment of Stage I and II Non-small Cell Lung Cancer: Diagnosis and Management of Lung Cancer, 3rd ed: American College of Chest Physicians Evidence-Based Clinical Practice Guidelines," Chest, vol. 143, no. 5, Supplement, pp. e278S-e313S, 2013.



\bibitem{Navani2018} N. Navani et al., "The Accuracy of Clinical Staging of Stage I-IIIa Non-Small Cell Lung Cancer: An Analysis Based on Individual Participant Data," Chest, vol. 155, no. 3, pp. 502-509, 2019.








\bibitem{Farjah2013}	F. Farjah, F. Lou, C. Sima, V. W. Rusch, and N. P. Rizk, "A Prediction Model for Pathologic N2 Disease in Lung Cancer Patients with a Negative Mediastinum by Positron Emission Tomography," Journal of Thoracic Oncology, vol. 8, no. 9, pp. 1170-1180, 2013.



\bibitem{Chen2013} K. Chen, F. Yang, G. Jiang, J. Li, and J. Wang, "Development and Validation of a Clinical Prediction Model for N2 Lymph Node Metastasis in Non-Small Cell Lung Cancer," The Annals of Thoracic Surgery, vol. 96, no. 5, pp. 1761-1768, 2013.






\bibitem{Gu2018}	Y. Gu, Y. She, D. Xie, C. Dai, Y. Ren, Z. Fan, H. Zhu et al., "A Texture Analysis–Based Prediction Model for Lymph Node Metastasis in Stage IA Lung Adenocarcinoma," The Annals of Thoracic Surgery, vol. 106, no. 1, pp. 214-220, 2018.

\bibitem{He2019}	L. He, Y. Huang, L. Yan, J. Zheng, C. Liang, and Z. Liu, "Radiomics-based predictive risk score: A scoring system for preoperatively predicting risk of lymph node metastasis in patients with resectable non-small cell lung cancer," Chinese journal of cancer research, vol. 31, no. 4, pp. 641-652, 2019.

\bibitem{Wang2019}	X. Wang, X. Zhao, Q. Li, W. Xia, Z. Peng, R. Zhang, Q. Li et al., "Can peritumoral radiomics increase the efficiency of the prediction for lymph node metastasis in clinical stage T1 lung adenocarcinoma on CT?," European Radiology, vol. 29, no. 11, pp. 6049-6058, 2019.








\bibitem{Wang2018} X. Wang, W. Nan, S. Yan, Q. Li, N. Guo, and Z. Guo, "MA05.11 Radiomics Analysis Using SVM Predicts Mediastinal Lymph Nodes Status of Squamous Cell Lung Cancer by Pre-Treatment Chest CT Scan," Journal of Thoracic Oncology, vol. 13, no. 10, pp. S374-S374, 2018.

\bibitem{Cong2020}	M. Cong, H. Feng, J.-L. Ren, Q. Xu, L. Cong, Z. Hou, Y.-y. Wang et al., "Development of a predictive radiomics model for lymph node metastases in pre-surgical CT-based stage IA non-small cell lung cancer," Lung Cancer, vol. 139, pp. 73-79, 2020.

\bibitem{Yoo2020}	J. Yoo, M. Cheon, Y. J. Park, S. H. Hyun, J. I. Zo, S.-W. Um, H.-H. Won et al., "Machine learning-based diagnostic method of pre-therapeutic 18F-FDG PET/CT for evaluating mediastinal lymph nodes in non-small cell lung cancer," European Radiology, 2020.

\bibitem{Hu2022} D. Hu, S. Li, H. Zhang, N. Wu, and X. Lu, "Using Natural Language Processing and Machine Learning to Preoperatively Predict Lymph Node Metastasis for Non–Small Cell Lung Cancer With Electronic Medical Records: Development and Validation Study," JMIR Medical Informatics, vol. 10, no. 4, p. e35475, 2022.








\bibitem{Zhao2020}	X. Zhao, X. Wang, W. Xia, Q. Li, L. Zhou, Q. Li, R. Zhang et al., "A cross-modal 3D deep learning for accurate lymph node metastasis prediction in clinical stage T1 lung adenocarcinoma," Lung Cancer, vol. 145, pp. 10-17, 2020.

\bibitem{Wang2017}	H. Wang, Z. Zhou, Y. Li, Z. Chen, P. Lu, W. Wang, W. Liu et al., "Comparison of machine learning methods for classifying mediastinal lymph node metastasis of non-small cell lung cancer from 18F-FDG PET/CT images," EJNMMI Research, vol. 7, no. 1, pp. 11-11, 2017.


\bibitem{Wang2021}	Y.-W. Wang, C.-J. Chen, H.-C. Huang, T.-C. Wang, H.-M. Chen, J.-Y. Shih, J.-S. Chen et al., "Dual energy CT image prediction on primary tumor of lung cancer for nodal metastasis using deep learning," Computerized Medical Imaging and Graphics, vol. 91, p. 101935, 2021.


\bibitem{Hu2023} D. Hu, S. Li, N. Wu, and X. Lu, "A Multi-modal Heterogeneous Graph Forest to Predict Lymph Node Metastasis of Non-small Cell Lung Cancer," IEEE Journal of Biomedical and Health Informatics, pp. 1-10, 2022.


\bibitem{Hu2024} D. Hu, B. Liu, L. Cheng, R. Guo, J. Wang, X. Lu, and N. Wu, "A Deep Multi-Task Network to Learn Tumor Pathological Representations for Lymph Node Metastasis Prediction," Studies in health technology and informatics, vol. 310, pp. 906-910, 2024.











\bibitem{OpenAI2024}	OpenAI. "Introducing ChatGPT." https://openai.com/blog/chatgpt (accessed March, 2024).

\bibitem{Achiam2023}	J. Achiam, S. Adler, S. Agarwal, L. Ahmad, I. Akkaya, F. L. Aleman, D. Almeida et al., "Gpt-4 technical report," arXiv preprint arXiv:2303.08774, 2023.













\bibitem{Brown2020}	T. Brown, B. Mann, N. Ryder, M. Subbiah, J. D. Kaplan, P. Dhariwal, A. Neelakantan et al., "Language Models are Few-Shot Learners," in Advances in Neural Information Processing Systems, H. Larochelle, M. Ranzato, R. Hadsell, M. F. Balcan, and H. Lin, Eds., 2020, vol. 33, pp. 1877--1901.







\bibitem{Ouyang2022}	L. Ouyang, J. Wu, X. Jiang, D. Almeida, C. Wainwright, P. Mishkin, C. Zhang et al., "Training language models to follow instructions with human feedback," in Advances in neural information processing systems, 2022, vol. 35, pp. 27730-27744. 









\bibitem{Tang2023}	L. Tang, Z. Sun, B. Idnay, J. G. Nestor, A. Soroush, P. A. Elias, Z. Xu et al., "Evaluating large language models on medical evidence summarization," npj Digital Medicine, vol. 6, no. 1, p. 158, 2023.



\bibitem{HuChen2024}	Y. Hu, Q. Chen, J. Du, X. Peng, V. K. Keloth, X. Zuo, Y. Zhou et al., "Improving large language models for clinical named entity recognition via prompt engineering," Journal of the American Medical Informatics Association, 2024.

\bibitem{Doshi2024}	R. Doshi, K. S. Amin, P. Khosla, S. Bajaj, S. Chheang, and H. P. Forman, "Quantitative Evaluation of Large Language Models to Streamline Radiology Report Impressions: A Multimodal Retrospective Analysis," Radiology, vol. 310, no. 3, p. e231593, 2024.





\bibitem{HuZhang2024}	D. Hu, S. Zhang, Q. Liu, X. Zhu, and B. Liu, "The current status of large language models in summarizing radiology report impressions," arXiv preprint arXiv:2406.02134, 2024.


\bibitem{HuLiu2023} D. Hu, B. Liu, X. Zhu, X. Lu, and N. Wu, "Zero-shot information extraction from radiological reports using ChatGPT," International Journal of Medical Informatics, vol. 183, p. 105321, 2024.





\bibitem{Chung2024}	P. Chung, C. T. Fong, A. M. Walters, N. Aghaeepour, M. Yetisgen, and V. N. O’Reilly-Shah, "Large Language Model Capabilities in Perioperative Risk Prediction and Prognostication," JAMA Surgery, 2024.


\bibitem{Glicksberg2024}	B. S. Glicksberg, P. Timsina, D. Patel, A. Sawant, A. Vaid, G. Raut, A. W. Charney et al., "Evaluating the accuracy of a state-of-the-art large language model for prediction of admissions from the emergency room," Journal of the American Medical Informatics Association, 2024.


\bibitem{Changho2024}	H. Changho, K. Dong Won, K. Songsoo, Seng, P. Jin Young, A. B. Sung, and Y. Dukyong, "Evaluation of GPT-4 for 10-year cardiovascular risk prediction: Insights from the UK Biobank and KoGES data," iScience, vol. 27, no. 2, p. 109022, 2024.


\bibitem{ZhuWang2024}	Y. Zhu, Z. Wang, J. Gao, Y. Tong, J. An, W. Liao, E. M. Harrison et al., "Prompting large language models for zero-shot clinical prediction with structured longitudinal electronic health record data," arXiv preprint arXiv:2402.01713, 2024.




\bibitem{HuZhangLi2021} D. Hu, H. Zhang, S. Li, Y. Wang, N. Wu, and X. Lu, "Automatic Extraction of Lung Cancer Staging Information From Computed Tomography Reports: Deep Learning Approach," JMIR Medical Informatics, vol. 9, no. 7, pp. e27955-e27955, 2021.













\bibitem{YanZhang2023}	Z. Yan, K. Zhang, R. Zhou, L. He, X. Li, and L. Sun, "Multimodal ChatGPT for medical applications: an experimental study of GPT-4V," arXiv preprint arXiv:2310.19061, 2023.



\bibitem{Nakao2024}	T. Nakao, S. Miki, Y. Nakamura, T. Kikuchi, Y. Nomura, S. Hanaoka, T. Yoshikawa et al., "Capability of GPT-4V(ision) in the Japanese National Medical Licensing Examination: Evaluation Study," (in English), JMIR Med Educ, Original Paper vol. 10, p. e54393, 2024.


\bibitem{ZhouOng2024}	Y. Zhou, H. Ong, P. Kennedy, C. C. Wu, J. Kazam, K. Hentel, A. Flanders et al., "Evaluating GPT-4V (GPT-4 with Vision) on Detection of Radiologic Findings on Chest Radiographs," Radiology, vol. 311, no. 2, p. e233270, 2024.

\bibitem{Brin2024}	D. Brin, V. Sorin, Y. Barash, E. Konen, G. Nadkarni, B. S. Glicksberg, and E. Klang, "Assessing GPT-4 Multimodal Performance in Radiological Image Analysis," medRxiv, p. 2023.11.15.23298583, 2024.






\end{thebibliography}
\end{document}